\documentclass[10pt,conference]{IEEEtran}
\usepackage{cite}
\usepackage{amsmath,amssymb,amsfonts}
\usepackage{thm-restate}
\usepackage{algorithmic}
\usepackage{graphicx}
\usepackage{textcomp}
\usepackage{xcolor}
\usepackage{xspace}
\usepackage{mathtools}
\usepackage{multirow, makecell}
\usepackage{tikz}
\allowdisplaybreaks

\ifCLASSOPTIONcompsoc 
\usepackage[caption=false,font=normalsize,labelfont=sf,textfont=sf]{subfig}
\else
\usepackage[caption=false,font=footnotesize]{subfig}
\fi

\newcommand\wrt{w.r.t.\xspace }
\newcommand\ie{i.e.,\xspace}
\newcommand\eg{e.g.,\xspace}
\newcommand\st{s.t.\xspace}
\newcommand\etc{etc.\xspace}
\newcommand\fig{Fig.\xspace}
\newcommand\reducespaceabovetbl{} 
\newcommand\reducespacebelowtbl{} 

\newcommand\set[2]{\{#1~|~#2\}}
\newcommand\setn[1]{\{#1\}}
\DeclareMathOperator*{\argmin}{arg\,min}

\renewcommand\emptyset\varnothing
\newcommand{\rom}[1]{\uppercase\expandafter{\romannumeral #1\relax}}
\newcommand{\mrow}[2]{\multirow{#1}{*}{\makecell[l]{#2}}}
\newcommand{\rtxt}[2]{\parbox[t]{1.5mm}{\mrow{#1}{\rotatebox[origin=c]{90}{#2}}}}

\DeclareFontFamily{OT1}{pzc}{}
\DeclareFontShape{OT1}{pzc}{m}{it}{<-> s * [1.10] pzcmi7t}{}
\DeclareMathAlphabet{\mathpzc}{OT1}{pzc}{m}{it}

\newcommand\optphi{\widehat{\phi}}

\newcommand\vseq{\mathcal{V}}
\newcommand\vseqn[1]{\mathpzc{v}^{(#1)}}

\newcommand\sg[1]{S(#1)}

\newcommand\allconfigs{\mathcal{C}}
\newcommand\costfn{\mathcal{F}}
\newcommand\reccostfn{\mathcal{R_\vseq}}
\newcommand\bfsreccostfn{\mathcal{B_\vseq}}

\newcommand\nodefn{\mathcal{H_\vseq}}

\newcommand\connected[1]{X(#1)}

\newcommand\dependent[1]{D(#1)}

\newcommand\bfsdependent[1]{D_B(#1)}

\newcommand\sortnodes{\textsc{GenerateSeq}\xspace}
\newcommand\dpalg{\textsc{FindBestStrategy}\xspace}
\newcommand\dfs{\textsc{DFS}\xspace}

\newcommand\mincost{min\_cost\xspace}

\newcommand{\RIGHTCOMMENT}[1]{\hfill\COMMENT{#1}}

\newcommand\copyrighttext{%
  \footnotesize \textcopyright 2021 IEEE. Personal use of this material is permitted.
  Permission from IEEE must be obtained for all other uses, in any current or future
  media, including reprinting/republishing this material for advertising or promotional
  purposes, creating new collective works, for resale or redistribution to servers or
  lists, or reuse of any copyrighted component of this work in other works.
  DOI: {10.1109/IPDPS49936.2021.00111}}
\newcommand\copyrightnotice{%
\begin{tikzpicture}[remember picture,overlay]
\node[anchor=south,yshift=10pt] at (current page.south) {\fbox{\parbox{\dimexpr\textwidth-\fboxsep-\fboxrule\relax}{\copyrighttext}}};
\end{tikzpicture}%
}

\begin{document}

\title{PaSE: Parallelization Strategies for Efficient DNN Training}

 \author{\IEEEauthorblockN{Venmugil Elango}
 \IEEEauthorblockA{\textit{Baidu Research, Sunnyvale, USA}\\
 Email: elango.4@buckeyemail.osu.edu}
 }

\maketitle
\copyrightnotice

\begin{abstract}
	Training a deep neural network (DNN) requires substantial computational and
memory requirements.  It is common to use multiple devices to train a DNN to
reduce the overall training time.  There are several choices to parallelize each
layer in a DNN.  Exhaustively searching this list to find an optimal
parallelization strategy is prohibitively time consuming and impractical.  The
standard practice is to use data parallelism because of its simplicity.
However, data parallelism is often sub-optimal, and suffers from poor
performance and high memory requirement. Expert-designed strategies have been
proposed on a case-by-case basis using domain specific knowledge.  These
expert-designed strategies do not generalize well to DNNs other than the ones
for which they were designed, and are not always necessarily the best choice.

In this paper, we propose an approach to automatically find efficient
parallelization strategies for DNNs from their computation graphs.  We present
an efficient algorithm to compute these strategies within a reasonable time in
practice.  We evaluate the effectiveness of our approach on various DNNs.  We
also compare the performance of the strategies identified by our approach
against data parallelism, expert-designed strategies, and the state-of-the-art
approaches.  Our results show that the strategies found using our approach
outperform the baseline data parallelism strategy in all the cases. In addition,
our strategies achieve better performance than the expert-designed
strategies and the state-of-the-art approaches.

\end{abstract}

\begin{IEEEkeywords}
Machine learning, neural nets, parallelism,
	automatic parallelization, dynamic programming, optimization.
\end{IEEEkeywords}

\section{\label{sec:intro}Introduction}

Deep neural networks are becoming increasingly sophisticated, and use
larger and larger datasets for better accuracies. This has led to an increase in
computational and memory requirements to train DNNs. It typically takes from
several hours to days, and multiple GPUs to train a network. For instance, as
noted in~\cite{gnmt}, Google's neural machine translation (GNMT) model takes
around 6 days to train on WMT EN$\rightarrow$FR dataset with 96 NVIDIA K80 GPUs.
Training a DNN involves three phases: \emph{forward propagation}, \emph{backward
propagation} (or \emph{backprop}), and \emph{update} phase.
First, the input dataset is split into multiple \emph{mini-batches}.
During a \emph{step}, a mini-batch is passed through the layers of the network
during forward propagation. At the end of the forward phase, the output is
compared against the \emph{ground truth}, and a \emph{loss} is computed
using an appropriate loss function. To minimize the loss, its \emph{gradients}
\wrt the model parameters are computed during backprop. Finally, the
model parameters are
updated during the update phase using the computed gradients. This process
is repeated over several timesteps, called \emph{epochs}, until the required
accuracy is achieved.

DNN parallelization strategies can be broadly classified into three, namely,
\emph{data parallelism}, \emph{parameter parallelism}, and \emph{pipeline
parallelism}. 
A strategy that combines these three approaches to parallelize each layer
differently is often referred to as \emph{hybrid parallelism}.
Each has its own advantages and disadvantages, as described below.

In \emph{data parallelism}, each of the $p$ devices keeps a replica of the entire
DNN, and a mini-batch is split into $p$ shards and distributed to different
devices. Each device performs forward and backward propagation independently on
its shard of data. During the update phase, gradients from all the devices are
accumulated, typically through an \emph{all-reduce} operation.
For large models, this communication becomes a major bottleneck. 
Further, as the model parameters are replicated (instead of
being distributed), it might be impossible to train large models by just
using data parallelism, due to memory constraints. Additionally, data parallelism
is inefficient at small mini-batch sizes. Unfortunately, using a larger
mini-batch size may not always be possible, owing to poor convergence and
accuracy~\cite{KeskarMNST16}. Despite these drawbacks, data parallelism remains
popular due to its simplicity.

An alternative strategy is to divide the work along model parameter and
attribute dimensions (\eg image height/width, channels, filters, \etc),
rather than the mini-batch
dimension. This is the approach taken by \emph{parameter
parallelism}\footnote{Some previous works~\cite{mesh-tf,
model-parallelism, owt} refer to this strategy as model
parallelism, while others~\cite{pipedream,gnmt} use the term model parallelism
to refer to a different strategy. To avoid any confusion, we instead use
the term parameter parallelism here. In terms of SOAP~\cite{flexflow},
parameter parallelism captures both attribute (A) and parameter (P)
dimensions. Refer to \fig~\ref{fig:gemm-config} for an illustration.} strategy~\cite{mesh-tf}. 
With this approach, the model parameters/attributes are distributed among different
devices, and each device only computes a part of a layer's activations (and
gradients) during forward (and backward) propagation.
This strategy typically incurs \emph{all-to-all} communication to accumulate
the activations and gradients. Depending on the mini-batch and model parameter
sizes, one strategy is more efficient than the other.

The third approach (\emph{pipeline parallelism})~\cite{gpipe} is to place
different layers of a network on different devices, without splitting the input
data or model parameters along any dimension. 
Each device computes activations (and gradients) for the layers it owns, and
sends the results to the devices that own the successive layers. This strategy
has the advantage of not needing to collectively communicate the model
parameters, however, there needs to be sufficient inter-layer parallelism and
the data needs to arrive at a specific rate through the pipeline for this
strategy to be efficient.

A parallelization strategy that combines multiple strategies to parallelize
different layers differently is typically referred to as \emph{hybrid
parallelism}~\cite{flexflow}. In hybrid parallelism, each layer is parallelized
differently using a mix of different strategies (\eg data+parameter
parallelism).  There are several possibilities 
to choose how different layers need to be parallelized.
Hence, it is impractical to exhaustively search for an optimal strategy.
Based on domain specific knowledge, expert designed
strategies~\cite{owt, gnmt} have been proposed on a case-by-case basis for
different DNNs.
There also have been efforts in the past to automatically find good strategies.
These works either (i) apply different heuristics~\cite{flexflow, reinforce} to
find a greedy solution, (ii) find an optimal solution restricted to
a certain class of DNNs~\cite{opt-cnn} (such as CNNs), or (iii)
reduce the search space by restricting some choices to find an optimal strategy
within the reduced search space~\cite{opt-cnn, reinforce, pipedream}.
In this paper, we take this third approach. We ignore \emph{inter-layer} pipeline
parallelism, and restrict ourselves to finding the best strategy to parallelize
different layers of a DNN using a combination of parameter and data parallelism. In
Section~\ref{sec:expt}, we empirically show that our method works well in
practice despite this restriction as it does not extensively prune the optimal
strategies from the search space. We also compare our results against the
state-of-the-art approach FlexFlow~\cite{flexflow}.  We formally define the
problem in Section~\ref{sec:representation}, and provide a method to find
efficient strategies in Section~\ref{sec:approach}.  In
Section~\ref{sec:related}, we summarize the previous works and discuss the
differences between our approach and theirs.
To summarize our contributions,
   \begin{itemize} 
		 \item We propose a formulation, and a vertex ordering strategy to enable
			 efficient computation of the best parallelization strategies for DNNs.
     \item We develop an efficient
       algorithm based on our
       formulation to compute the best strategies for various DNNs.
       A prototype implementation of our approach is available at
       https://github.com/baidu-research/PaSE.
       Experimental results show that our algorithm finds efficient strategies
       within a few seconds for various DNNs.
		 \item We evaluate the strategies found by our approach against
			 data parallelism, expert-designed strategies, and the strategies proposed
			 by the state-of-the-art framework FlexFlow~\cite{flexflow}.
			 Results show that our strategies outperform data parallelism by up
			 to $1.85\times$ on a multi-node/multi-GPU system consisting of 1080Ti
			 GPUs, and by up to $4\times$ on a system consisting of 2080Ti GPUs for
			 various benchmarks. Our strategies also perform better than
			 expert-designed strategies, and the strategies suggested by FlexFlow.
	 \end{itemize}

\section{\label{sec:representation}Problem representation}
%
A DNN can be represented as a \emph{computation graph}. A computation graph
$G=(V,E)$ is a weakly connected directed graph, where each node $v\in V$
corresponds to a layer (\eg fully-connected, convolution, \etc) in the DNN, and
each edge $(u, v)\in E$ represents flow of a tensor that is an output of $u$ and
an input of $v$.
Each node $v\in V$ has an associated \emph{iteration space}~\cite{tiling}
that captures the computation of $v$.
Consider, for instance, a fully-connected layer that multiplies a matrix
$A_{M\times K}$ with a matrix $B_{K\times N}$. Its iteration space is
specified by the set $\set{(i, j, k) \in \mathbb{Z}^3}{0 \le i < M \land 0 \le j
< N \land 0 \le k < K }$.

\begin{figure}[!htbp]
	\centering
	\includegraphics[width=.23\textwidth,trim={0 0 0 .7cm},clip]{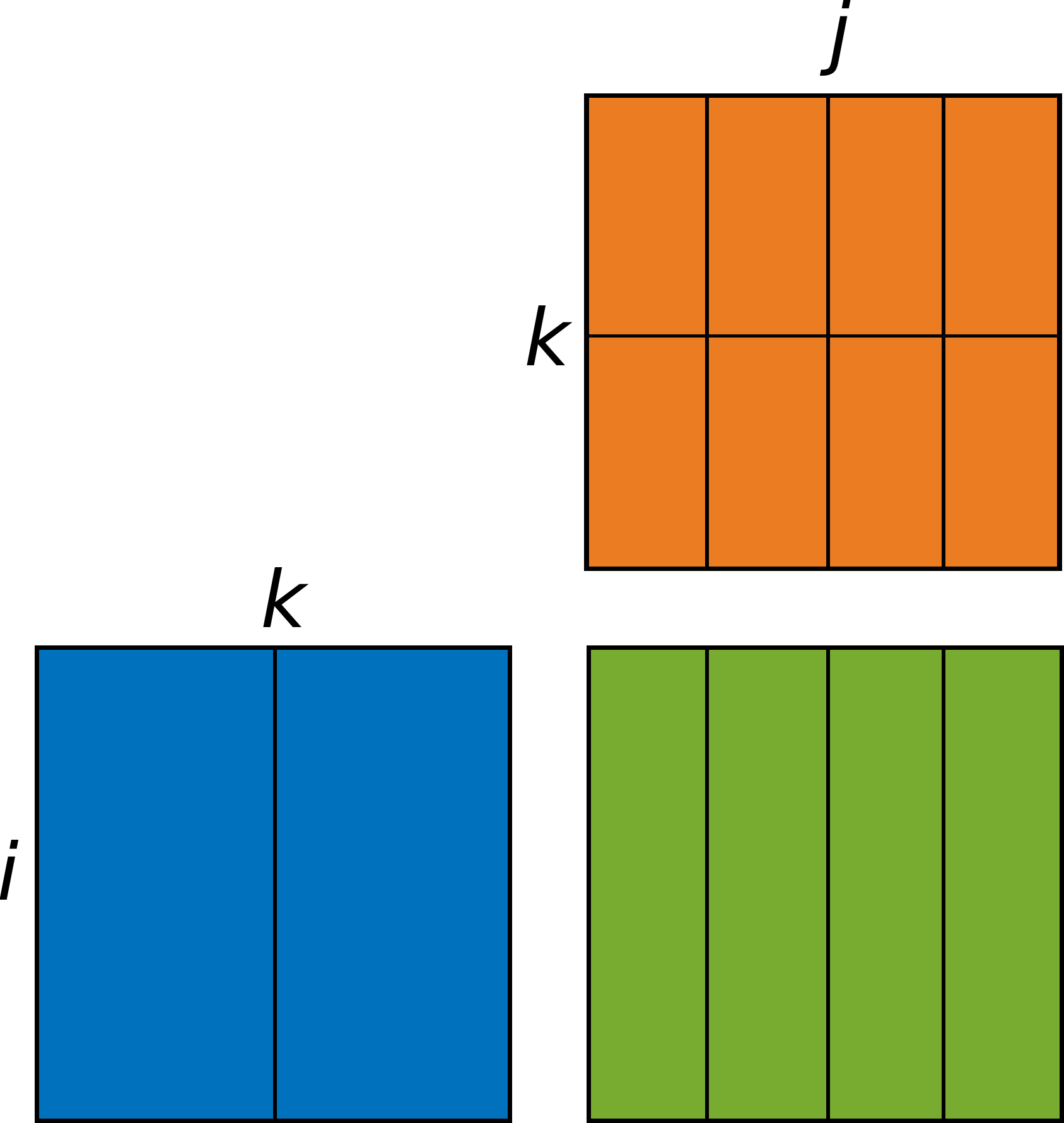}
	\caption{\label{fig:gemm-config} Iteration space of a GEMM computation
	parallelized using the configuration $(1, 4, 2)$. $j$ and $k$ dimensions are
	split $4$-ways and $2$-ways, respectively, while the $i$ dimension is not
	parallelized.}
\end{figure}

A \emph{parallelization configuration}~\cite{opt-cnn} $C_v$ of a node $v$ is a
\mbox{$d$-tuple} of positive integers that defines how the iteration space of $v$
is split and
parallelized across different devices, where $d$ is the dimension of the
iteration space of $v$. For the fully-connected layer example above, a
configuration $(1, 4, 2)$ states that the iteration space has to be split into $4$
equal parts along the second dimension of its iteration space, and into $2$
parts along the third dimension. (Refer \fig~\ref{fig:gemm-config}.)
Computationally, this configuration states: split columns of $A$ and
rows of $B$ into two equal parts; split columns of $B$ into four equal parts;
perform the 8 GEMM computations that correspond to each part on 8 separate
devices; and finally perform partial reduction of the intermediate results.
Refer~\cite[Section 6]{mesh-tf} for a more concrete description of how
different DNN layer types can be parallelized, and the communications they
incur.
Given a node $v$ with $d$ dimensional
iteration space, and $p$ devices, the set of valid configurations for $v$
is given by $\allconfigs(v) = \set{(c_1,\dots,c_d)\in
\mathbb{N}^d}{\prod_{i=1}^d c_i \le p}$. 

A \emph{parallelization strategy} $\phi$ is the set $\set{(v, C_v)}{v\in V \land
C_v\in \allconfigs(v)}$ that specifies a valid configuration for each node $v\in
V$. Configuration for a node $v$ in $\phi$ is given by $C_v=\phi(v)$.
A \emph{substrategy} $\phi_{|U}$ is a strategy restricted to
the subset $U$, \ie $\phi_{|U} = \set{(u, \phi(u))}{u\in U}$.
An \emph{optimal strategy} $\optphi$ has the minimum cost over all possible
strategies for $V$, under a given cost function $\costfn$, \ie $\optphi
= \argmin_{\phi\in \Phi} \costfn(G, \phi)$, where $\Phi$ is the set of all
valid strategies for $V$.
%
Given $p$ devices with an average peak floating-point performance of $F$ FLOPS
per device, and an average communication bandwidth of $B$ bytes per second per
link, the cost function we use is:
\begin{equation}\label{eq:costfn}
\costfn(G, \phi) =  \sum_{v\in V} t_l(v, \phi, r)  
	 + \sum_{(u,v)\in E} r\times t_x(u, v, \phi)
\end{equation}
where, $r = F / B$ is the FLOP-to-bytes ratio; layer cost $t_l(v, \phi, r)$
is the cost (in FLOP) of executing the layer $v$ (such as fully-connected)
using the parallelization configuration $\phi(v)$  -- this
cost includes both
computation and communication that happens internally within a layer (such as
all-reduce within a layer, halo communication for convolutions, \etc, normalized to FLOP by multiplying it
with $r$); and data transfer cost $t_x(u, v, \phi)$ is the communication cost (in bytes)
needed to communicate the tensor that flows along the edge $(u, v)$ or $(v, u)$
during forward and backward propagation, where $u$ and $v$ are parallelized
using configurations $\phi(u)$ and $\phi(v)$, respectively.\footnote{Note that $t_x$ captures
communication cost along both directions (forward and backward), and is
edge-direction agnostic, \ie for an edge $(u, v)\in E$, $t_x(u, v, \phi)
= t_x(v, u, \phi)$.}

Our cost function	$\costfn$ is an approximation of the actual cost. It ignores
any overlapping (or pipelining) of different layers by adding the costs
$t_l(v_x,\cdot,\cdot)$ and $t_l(v_y,\cdot,\cdot)$ of any two layers (instead of
taking a $\max$ where possible).
Thus, it captures data and parameter
parallelism, but ignores \emph{inter-layer} pipeline parallelism. 
Note that this only ignores
pipeline parallelism between layers, while any \emph{intra-layer} pipeline parallelism
opportunities within a layer can be accurately captured by 
accounting for it in the layer cost $t_l$.
In return, this approximation allows us to devise a technique (described in
Section~\ref{sec:approach}) to efficiently find the best strategies quickly in
practice.
Our approach is effective despite this simplification as most DNNs do not
contain significant inherent pipeline parallelism opportunities due to data
dependence constraints.
Even though our approach finds the
optimal solution $\optphi = \argmin_{\phi\in \Phi} \costfn(G, \phi)$ under the
cost function
$\costfn$, since $\costfn$ itself is an approximation, rather
than referring to our solution as the \emph{optimal} strategy, we refer
to it as an \emph{efficient} strategy or the \emph{best} strategy, to avoid any
confusion. 
Some previous works \cite{pipedream} have used
\emph{inter-batch pipeline parallelism} to improve parallel training throughput
by making semantic modifications to the model. 
In this work, we do not perform any such semantic modifications to the model.
Thus, the convergence rates and the final accuracies of the strategies proposed
by our method are exactly same as the original model.


As we only have a handful of different types of DNN layers, we use analytically
derived layer costs $t_l$ (parametrized for problem sizes) for different types
of layers.
Communication cost $t_x$ along an edge $(u, v)$ is given by:
\textsc{$\max_d |A(v,d,\phi)| - |A(v,d,\phi)\cap A(u,d,\phi)|$},
where, $A(v,d,\phi)$ and $A(u,d,\phi)$ are the volume of input tensor of $v$
(parallelized using configuration $\phi(v)$) needed by a
device $d$, and volume of output tensor of $u$ (parallelized using $\phi(u)$)
held by $d$, respectively.
We ignore many low level details such as cache effects, \etc, to keep the cost
function simple, although such simplifications are not necessary for our method
to work.
Additionally, to select the best strategy, our approach only requires various strategies to be
ranked in the correct order. For instance, if a strategy $\phi_1$ is better than
$\phi_2$, the analytical cost of $\phi_1$ computed by our cost function
($\costfn$) has to be lower than $\phi_2$. However, precisely predicting their
absolute runtime costs is not necessary for our method to work accurately.
Our simplifying assumptions affect costs of all the strategies more or less
alike, preserving most of the relative ordering.

Although a parallelization configuration only describes how to split
an iteration space into multiple parts, while the actual device assignment for
each part is not explicitly specified, our experience shows
that once we have a complete parallelization strategy, a simple
greedy assignment that maximizes data locality 
(\ie a greedy assignment
that maximizes $|A(v,d,\phi)\cap A(u,d,\phi)|$)
works sufficiently well in practice. Additionally, frameworks such as
GShard~\cite{gshard} can take user-specified parallelization strategies, such
as the ones computed by our approach, and automatically perform efficient device
assignment by simply aligning the sharding decisions of adjacent layers.

Even though our objective primarily focuses on 
minimizing the training time, this also indirectly 
minimizes the space requirements, since
the memory footprint per device is a sum of (i) the space needed to
hold the input and output tensors, and
(ii) the space for the communication buffers. 
When a DNN layer is distributed among $d$ devices,
the space required for (i) often reduces by a factor proportional
to $d$ uniformly for all parallelization strategies (there are a few
uncommon cases where this may not strictly hold), and the space required
by (ii) is proportional to the amount of
communication, which our objective indeed tries to minimize.

\section{\label{sec:approach}Computing Efficient Strategies}
This section describes our approach to compute the best strategies for DNNs
efficiently.

\paragraph*{Notation}
Given $G=(V,E)$ and a vertex $v\in V$, we let
$N(v)$ denote its neighbors, \ie $N(v) = \set{u \in V}{(u, v)\in E \lor (v, u)
\in E}$. Given some $U\subseteq V$, we also use the notation $N(U)
= \bigcup_{u\in U}N(u)$ to refer to the neighbors of $U$.
Let $\vseq=(\vseqn{1},\dots,\vseqn{|V|})$ be an (arbitrary) ordering of $V$.
Then $\vseqn{i}$ refers to the $i\textsuperscript{th}$
vertex in the sequence $\vseq$.
We also use $\vseq_{\le i}$, $\vseq_{\ge i}$, $\vseq_{<i}$, and $\vseq_{>i}$ to
refer to vertex sets $\setn{\vseqn{1},\dots,\vseqn{i}}$,
$\setn{\vseqn{i},\dots,\vseqn{|V|}}$, $\setn{\vseqn{1},\dots,\vseqn{i-1}}$ and
$\setn{\vseqn{i+1},\dots,\vseqn{|V|}}$, respectively.


\subsection{\label{subsec:bfs}A na\"ive approach}
A brute-force method to find an efficient strategy for a computation graph
is to enumerate all possible combinations of configurations of the
vertices, and choose the one with the least cost.
Combinatorial nature of this method makes it impractical to use even on
small graphs such as AlexNet.
However, a straight-forward observation shows that the parallelization
configuration chosen for a layer only affects the cost of computing the layer
itself, and its neighbors. This is also evident from
Equation~(\ref{eq:costfn}), where, changing the configuration for a vertex $v$
from $C$ to $C'$ only affects the layer cost $t_l(v,\cdot,\cdot)$ of the
vertex itself, and its data transfer costs with its neighbors
$t_x(u,v,\cdot)$, where $u\in N(v)$.
One way to exploit this property is to sequence $V$ in
a \emph{breadth-first} traversal order $\vseq=(\vseqn{1},\dots,\vseqn{|V|})$,
and find the best strategy for $G$ as follows:
For an $i^\textsuperscript{th}$ vertex $\vseqn{i}$ in $\vseq$, we define
its \emph{dependent set} $\bfsdependent{i}$ as the set of neighbors of
$\vseq_{\le i}$ that are in $\vseq_{>i}$, \ie $\bfsdependent{i}=N(\vseq_{\le
i})\cap \vseq_{>i}$.
Let $\phi\in \Phi_{|\bfsdependent{i}}$ be any valid substrategy for
$\bfsdependent{i}$.
Then, for $1\le i\le |V|$, a configuration $C$ for $\vseqn{i}$ that minimizes
the overall parallel training cost is given by the following recurrence:
\begin{equation}\label{eq:bfs-recurrence}
	\begin{split}
    \bfsreccostfn(i, \phi)  ={}& \min_{C\in \allconfigs(\vseqn{i})}~
    \nodefn(i, \phi') + \bfsreccostfn(i-1, \phi'') \\
		\bfsreccostfn(0, \phi)  = {}&0
	\end{split}
\end{equation}
where, $\phi' = \phi \cup \setn{(\vseqn{i},C)}$, $\phi''
= \phi'_{|\bfsdependent{i-1}}$,
and
\begin{equation}\label{eq:nodefn}
	\nodefn(i, \phi) = t_l(\vseqn{i}, \phi, r)   +
  \sum_{v \in N(\vseqn{i})\cap \vseq_{>i}}
	r \times t_x(\vseqn{i}, v, \phi)
\end{equation}
The function $\nodefn(i, \cdot)$ captures the layer cost of $\vseqn{i}$ and its
data transfer cost with its neighbors that appear after $\vseqn{i}$ in $\vseq$.
(Data transfer costs with its remaining neighbors are captured within
$\bfsreccostfn(i-1,\cdot)$.)
An efficient parallelization strategy for $G$ is the one that achieves the
minimum cost $\bfsreccostfn(|V|, \emptyset)$.
As the recurrence (\ref{eq:bfs-recurrence}) has an optimal substructure, a
dynamic programming (DP) based algorithm can be used to find the best strategy.
However, as we show later in Table~\ref{tbl:running-times}, computing efficient
strategies using this recurrence is still quite expensive, and takes significant
amount of time to find the best strategies for graphs other than simple path
graphs such as AlexNet. In the next subsection, we will derive a more efficient
approach to find the best strategies.

\subsection{\label{subsec:sortnodes}An efficient approach}
From recurrence~(\ref{eq:bfs-recurrence}), we can observe that as
$\bfsreccostfn(i, \phi)$ needs to be computed for
all possible $\phi\in \Phi_{|\bfsdependent{i}}$,
the computational
complexity for finding the best strategy using
(\ref{eq:bfs-recurrence}) is at least $O(K^{M+1})$, where $K=\max_{v\in
V}|\allconfigs(v)|$ is the maximum number of configurations for any vertex in
$G$, and $M=\max_{i\in[1,|V|]} |\bfsdependent{i}|$ is the size of the largest
dependent set.
It is important to note that dependent sets (and thus $M$) are a function of
sequence $\vseq$. Thus, by carefully arranging the vertices in $\vseq$, it
is possible to reduce $M$ and thus the overall computational complexity. 

\begin{figure}[!htbp]
	\centering
  \subfloat[\label{fig:toy-graph} $G=(V,E)$ ]
	{\includegraphics[width=5cm]{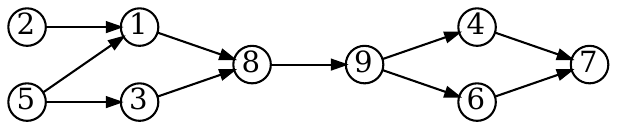}}
  \\
  \subfloat[\label{fig:toy-seq} $\vseq$: an ordering of $V$ from left to right.
  ($\vseq$ is edge direction agnostic. Edges can go either forward or backward.) ]
	{\includegraphics[width=7cm]{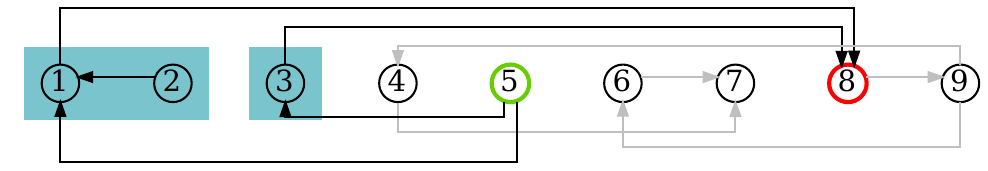}}
	\caption{\label{fig:toy-graph-seq}A toy computation graph $G$, and an ordering
  $\vseq$ of its vertices. 
  For the vertex $\vseqn{5}$ (marked in green), its connected set
  $\connected{5}=\setn{\vseqn{1}, \vseqn{2}, \vseqn{3}, \vseqn{5}}$, and its
  dependent set $\dependent{5}=\setn{\vseqn{8}}$ (marked in red). Its
  connected subsets $\sg{i}=\setn{\setn{\vseqn{1},\vseqn{2}}, \setn{\vseqn{3}}}$
  are represented by blue boxes in the figure.
	A similar, but more elaborate, structure appears in
  InceptionV3 (refer \fig~\ref{fig:inception-graph}) and Transformer models.}
\end{figure}

If computation graphs are fully sparse, any arbitrary ordering would be adequate to
keep $M$ sufficiently small, whereas if they are fully dense, no possible
ordering can help reduce $M$. However, 
a unique property of DNN graphs is that they are mostly
sparse with a few high degree nodes. Thus, if an arbitrary ordering, or
a simple breadth-first ordering is used, these dense locations form
the bottlenecks, leading to high computational overhead in finding the best
strategies. 
By carefully ordering the vertices, the size $M$ can be reduced, allowing
us to compute the parallelization strategies efficiently.
In this subsection, we develop an algorithm that exploits this property to
order the vertices in a sequence that keeps the sizes of these dependent sets to
the minimum.
We first introduce a few definitions and reformulate the recurrence
(\ref{eq:bfs-recurrence}) in terms of subgraphs and connected components
that provides us the flexibility to efficiently compute costs from arbitrary
sequences.
Let $G=(V,E)$ be the computation graph for a DNN, and $\vseq$ be an ordering of
$V$. For a vertex $\vseqn{i}$:
\paragraph{Connected set}
A \emph{connected set} $\connected{i}\subseteq \vseq_{\le i}$ of
$\vseqn{i}$ is the set of vertices in $\vseq_{\le i}$ that are connected to
$\vseqn{i}$ through a path $(\mathmbox{v_1\in\vseq_{\le i}},\dots)$
that only goes through the vertices in $\vseq_{\le i}$.
Note that $\vseqn{i}\in \connected{i}$.
For \eg in
\fig~\ref{fig:toy-graph-seq}, $\connected{5}=\setn{\vseqn{1}, \vseqn{2},
\vseqn{3}, \vseqn{5}}$.
Intuitively, any vertex $v\not\in \connected{i}$ is irrelevant, both directly
and indirectly, for finding the best configuration for $\vseqn{i}$, and thus
can be ignored allowing us to use a smaller dependent set as defined below.
\paragraph{Dependent set}
We redefine the \emph{dependent set} of $\vseqn{i}$ as the neighbors of
$\connected{i}$ that are in $\vseq_{> i}$, \ie
$\dependent{i}=N(\connected{i})\cap \vseq_{> i}$.
\paragraph{Connected subsets}
Given a connected set $\connected{i}$, consider the vertex set
$U=\connected{i}-\setn{\vseqn{i}}$, and its induced subgraph $G'=(U, E\cap
U\times U)$. $G'$ is composed of one or more connected components
$\setn{(V_1,E_1),\dots}$. We refer to the set of vertices of these connected
components $\setn{V_1,\dots}$ as the \emph{connected subsets} $\sg{i}$ of
$\vseqn{i}$. In \fig~\ref{fig:toy-graph-seq},
$\sg{5}=\setn{\setn{\vseqn{1},\vseqn{2}}, \setn{\vseqn{3}}}$.
Intuitively, the problem of finding the best configuration for $\vseqn{i}$ is
broken down into smaller sub-problems in terms of $\sg{i}$, that allows us to
use dynamic programming to solve our new recurrence (\ref{eq:recurrence})
defined below.

Let $\vseqn{i}$ be the
$i\textsuperscript{th}$ vertex in $\vseq$, and $\phi\in \Phi_{|\dependent{i}}$
be a valid substrategy for the vertices in $\dependent{i}$. Then,
\begin{equation}\label{eq:recurrence}
  \begin{split}
    \reccostfn(i, \phi) = \min_{C\in \allconfigs(\vseqn{i})}~
    \nodefn(i, \phi') + \sum_{\connected{j}\in \sg{i}}
    \reccostfn(j, \phi'')
  \end{split}
\end{equation}
where, $\phi' = \phi\cup \setn{(\vseqn{i},C)}$,
$\phi''=\phi'_{|\dependent{j}}$, and
$\nodefn$ is defined in (\ref{eq:nodefn}). We show in
Theorem~\ref{thm:optimality} below that a parallelization strategy
that minimizes $\reccostfn(|V|, \emptyset)$ corresponds to an efficient strategy
to parallelize $G$. We provide a proof for Theorem~\ref{thm:optimality} in
Appendix~\ref{subsec:optimality}.
\begin{restatable}{theorem}{optimality}\label{thm:optimality}
	Let $G=(V, E)$ be a computation graph for a DNN that is executed 
	on $p$ devices with average FLOP-to-bytes ratio $r$.
  Let $\vseq$ be a sequence for $V$, and $\Phi$ be the set of all possible
  strategies for $G$. 
  Then,
  \begin{equation*}
    \reccostfn(|V|, \emptyset) = \min_{\phi\in \Phi}~\costfn(G, \phi).
  \end{equation*}
\end{restatable}

Note that the recurrence (\ref{eq:recurrence}) by itself does not reduce the
complexity of finding the best strategies. For instance, with
a simple breadth-first ordering, for any vertex $\vseqn{i}$, $\connected{i}
= \vseq_{\le i}$, and thus $\dependent{i}= \bfsdependent{i}$.
However, 
recurrence (\ref{eq:recurrence}) provides the flexibility to efficiently
compute costs if the sequence $\vseq$ has the potential. 
For instance, in
\fig~\ref{fig:toy-graph-seq}, $|\bfsdependent{5}|=|\setn{\vseqn{7},
\vseqn{8}, \vseqn{9}}|=3$, while $|\dependent{5}|=|\setn{\vseqn{8}}|=1$. 
Since the computational complexity of the recurrence is exponential in terms of
the sizes of dependent sets, using recurrence (\ref{eq:recurrence}) instead of
(\ref{eq:bfs-recurrence}) exponentially reduces the computation time for
finding the best strategy. 

We will now develop an approach to generate a sequence $\vseq$ that will maintain
the sizes of dependent sets of the vertices as small as possible, thus allowing
us to efficiently compute the recurrence~(\ref{eq:recurrence}).
The complete algorithm \sortnodes is shown in \fig~\ref{alg:sortnodes}.
\sortnodes generates a sequence that maintains
the sizes of dependent sets $\dependent{i}$ (referred to as
$v.d$ in \fig~\ref{alg:sortnodes}) as small as possible.
\begin{figure}[!htbp]
	\begin{algorithmic}[1]
	\REQUIRE \sortnodes$(G=(V, E))$
    \STATE $\forall_{v\in V}, v.d \gets N(v)$ \label{ln:sort-init-d}
		\STATE $U \leftarrow V$ \RIGHTCOMMENT{Unsequenced nodes}\label{ln:u-init}
		\STATE $\vseq = (\bot_1,\dots,\bot_{|V|})$

		\FOR{$i=1$ \TO $|V|$} \label{ln:sort-begin-for}
		\STATE $\vseqn{i} \gets \argmin_{u \in U} |u.d|$ \RIGHTCOMMENT{Assign
		$i\textsuperscript{th}$\label{ln:sort-min-node}\label{ln:sort-set-v}
		element of $\vseq$}
		\STATE $U \gets U - \setn{\vseqn{i}}$
    \label{ln:sort-vx-update}

		\FORALL{$v \in \vseqn{i}.d$}\label{ln:sort-iter-vj}
		\STATE $v.d \gets v.d \cup \vseqn{i}.d - \setn{\vseqn{i}}$\label{ln:sort-vd}
		\ENDFOR \label{ln:sort-end-for}

		\ENDFOR
		\RETURN $\vseq$
	\end{algorithmic}
	\caption{\label{alg:sortnodes}Algorithm to generate a sequence $\vseq$ such that
	sizes of dependent sets are small.}
\end{figure}
In Line~\ref{ln:sort-init-d}, the dependent set of a vertex $v$ is initialized
to its neighbors.
In Line~\ref{ln:sort-min-node}, at an iteration $i$, a node $u$ that has the least
cardinality $|u.d|$ is picked from the set of nodes $U$ that are yet to be
sequenced. This node becomes $\vseqn{i}$ in $\vseq$.
Once $u$ has been added to $\vseq$, $v.d$ for all the nodes in $\vseqn{i}.d$ are
updated in Line~\ref{ln:sort-vd}. 
This makes sure that $|v.d|$ that is checked in Line~\ref{ln:sort-min-node} is
correctly maintained. 
%
%
%
In Theorem~\ref{thm:sort-correctness} (proof available in
Appendix~\ref{subsec:correctness}), we show that the dependent sets
computed by \sortnodes are indeed correct.  Computational complexity of
\sortnodes is $O({|V|}^2)$.
\begin{restatable}{theorem}{correctness} \label{thm:sort-correctness}
	Given a computation graph $G=(V, E)$, and a sequence $\vseq$ computed by
	\sortnodes in \fig~\ref{alg:sortnodes},
  for any $\vseqn{i}\in V$, $\vseqn{i}.d = \dependent{i}$.
\end{restatable}

A dynamic programming (DP) based algorithm for recurrence~(\ref{eq:recurrence})
that uses \sortnodes to compute an efficient strategy is shown in
\fig~\ref{alg:dp}.
In Line~\ref{ln:dp-sort}, $V$ is sequenced using \sortnodes. 
In Line~\ref{ln:dp-sub-strategies}, all possible valid substrategies $\Phi$
for the set $\dependent{i}$ are computed.
The function $\dfs(G,U,v)$ performs depth-first search on subgraph of $G$
induced by $U$, starting from the vertex $v$, to obtain the vertices
reachable from $v$ passing only through $U$. This function is used to compute
$\connected{i}$ and $\sg{i}$ in lines \ref{ln:dp-x} and \ref{ln:dp-s},
respectively.
For each $\phi\in
\Phi$, a configuration $C\in \allconfigs(\vseqn{i})$ that minimizes the cost
$\reccostfn(i, \phi\cup \setn{(\vseqn{i}, C)})$ is computed in
lines~\ref{ln:dp-c-loop-start}--\ref{ln:dp-c-loop-end}: Line~\ref{ln:dp-g}
computes $\nodefn(i, \phi')$;
lines~\ref{ln:dp-t-loop-start}--\ref{ln:dp-t-loop-end} use DP tables
$\vseqn{j}.tbl$ to get the costs
$\sum_{\connected{j}\in \sg{i}} \reccostfn(j, \phi'')$.
For a substrategy $\phi$, if a better configuration $C$ is found for
$\vseqn{i}$, $\vseqn{i}.cfg$ and $\vseqn{i}.tbl$ are updated in
lines~\ref{ln:dp-Phi-update-tbl}--\ref{ln:dp-Phi-update-cfg}.
Finally, in Line~\ref{ln:dp-end}, the minimum cost of the best strategy for $G$
is returned.
For simplicity, we do not show the details of extracting the best strategy from
the stored configurations $\vseqn{i}.cfg$, but a simple back-substitution,
starting from $\vseqn{|V|}.cfg$ provides the best strategy found by \dpalg.
\begin{figure}[!htbp]
	\begin{algorithmic}[1]
	\REQUIRE \dpalg$(G=(V, E))$

		\STATE $\vseq \gets \sortnodes(G)$ \RIGHTCOMMENT{Refer
		\fig~\ref{alg:sortnodes}}\label{ln:dp-sort}
    \STATE $\forall_{v\in V}, v.tbl \gets \emptyset$\RIGHTCOMMENT{DP table}
    \STATE $\forall_{v\in V}, v.cfg \gets \emptyset$\RIGHTCOMMENT{Best configs}

		\FOR{$i=1$ \TO $|V|$}\label{ln:dp-main-loop-start}
    \STATE $\Phi \gets \prod_{v\in \vseqn{i}.d} \set{(v, C)}{C\in
    \allconfigs(v)}$\RIGHTCOMMENT{$\Phi_{|\dependent{i}}$}\label{ln:dp-sub-strategies}
    \STATE $X \gets \dfs(G, \vseq_{\le i}, \vseqn{i})$
    \RIGHTCOMMENT{\connected{i}} \label{ln:dp-x}
    \STATE $S \gets \bigcup_{v\in X
    - \setn{\vseqn{i}}}\dfs(G, \vseq_{< i}, v)$
    \RIGHTCOMMENT{\sg{i}} \label{ln:dp-s}

    \FORALL{$\phi\in \Phi \lor \setn{\emptyset}$}\label{ln:dp-phi-loop-start}
		\STATE $\mincost \gets \top$

		\FORALL{$C \in \allconfigs(\vseqn{i})$}\label{ln:dp-c-loop-start}
		\STATE $\phi' \gets \phi \cup \setn{(\vseqn{i}, C)}$
		\STATE $cost \gets \nodefn(i, \phi')$\label{ln:dp-g}

		\FORALL{$X' \in S$}\label{ln:dp-t-loop-start}
    \STATE $j\gets \max_{\vseqn{k}\in X'}~k$
    \STATE $\phi'' \gets \set{(v,\phi'(v))}{v\in \vseqn{j}.d}$
    \STATE $cost \gets cost + \vseqn{j}.tbl(\phi'')$\label{ln:dp-sum-d}
		\ENDFOR\label{ln:dp-t-loop-end}

		\IF{$cost < \mincost$}
    \STATE $\vseqn{i}.tbl(\phi) \gets \mincost \gets cost$\label{ln:dp-Phi-update-tbl}
    \STATE $\vseqn{i}.cfg(\phi) \gets C$\label{ln:dp-Phi-update-cfg}
		\ENDIF
		\ENDFOR\label{ln:dp-c-loop-end}
		\ENDFOR\label{ln:dp-phi-loop-end}
		\ENDFOR

    \RETURN $\vseqn{|V|}.tbl(\emptyset)$\label{ln:dp-end}

	\end{algorithmic}
	\caption{\label{alg:dp}Dynamic programming based algorithm to compute an
	efficient strategy for a computation graph $G$.}
\end{figure}
Overall computational complexity of \dpalg is $O({|V|}^2K^{M+1})$, where
$K=\max_{v\in V} |\allconfigs(v)|$ is the maximum number of configurations for a
layer, and $M=\max_{\vseqn{i}\in V} |\dependent{i}|$ is the size of the
largest dependent set.

%

\subsection{\label{subsec:eg-inception}Example: InceptionV3}
\begin{figure}[!htbp]
	\centering
  \includegraphics[width=.78\linewidth,trim={233cm 0.42cm 27cm 1.9cm},clip]{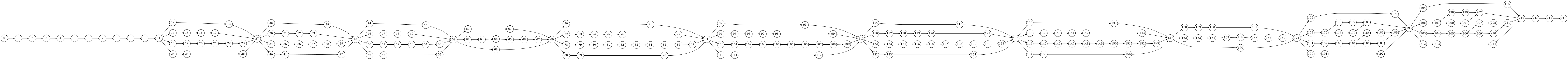}
	\caption{\label{fig:inception-graph}Computation subgraph corresponding to
	InceptionE module of InceptionV3. A similar structure repeats
	throughout the graph. Nodes $171$ and $193$ have high degree, while the rest
	of the nodes are sparse.}
\end{figure}

As described earlier, DNN graphs are generally sparse with a few high
degree nodes.
For instance, the computation graph of InceptionV3
(\fig~\ref{fig:inception-graph}) has $218$ nodes, of which $206$ of
them have a node degree of $<5$ and the remaining $12$ nodes have degree $\ge
5$.
Number of parallelization configurations per vertex of InceptionV3 vary between
$10$ and $30$ for $p=8$ GPUs, and the maximum number of configurations reaches
up to 100 (\ie $K=100$) for $p=64$ GPUs.
Our experiments show that when breadth-first ordering is used, the sizes of
dependent sets reach up to $10$, leading to $K^{M+1}\ge {10}^{11}$ (for $p=8$)
combinations to be analyzed to find the best configuration, making it
prohibitively expensive in practice, in terms of both time and space (Refer
Table~\ref{tbl:running-times}).
However, by ordering the vertices using \sortnodes,
$|\dependent{i}\cup\setn{\vseqn{i}}|$ for any $i$ is maintained to be $\le 3$,
and the maximum number of combinations analyzed per vertex by the algorithm,
$K^{M+1} \le 25200$, for $p=8$, enabling us to find the best configurations within
a few seconds.
This is because \sortnodes makes sure that the nodes with
high degree (nodes $171$ and $193$ in \fig~\ref{fig:inception-graph}) are
placed in the sequence only after their (low degree) neighbors, and their
ancestors/descendants are placed in the sequence, thus ensuring that the
sizes of dependent sets of these high degree nodes remains small.
%
%
%

\section{\label{sec:expt}Experimental Results}
We evaluate our technique on four different benchmarks, each having a different
graph structure:
a)~AlexNet~\cite{alexnet} is a image
classification \emph{convolutional network} whose computation graph is a simple
path graph, where each layer is only connected to the next layer;
b)~InceptionV3~\cite{inception} is a \emph{deep CNN} that uses inception modules to
increase the number of layers while maintaining a reasonable computational
budget. The nodes are split and concatenated (refer
\fig~\ref{fig:inception-graph}) at the beginning and end of each
inception module, respectively, leading to a few high degree nodes;
c)~RNNLM~\cite{rnnlm} is a two-layer \emph{recurrent} neural network consisting
of LSTM cells, used for \emph{language modeling} tasks; and finally,
d)~Transformer~\cite{transformer} model is a \emph{non-recurrent neural machine
translation} model, whose computation graph is quite different from recurrent
networks such as RNNLM.
We used ImageNet-1K~\cite{imagenet} dataset for CNNs,
Billion-Word~\cite{billion-word} for RNNLM, and WMT EN$\rightarrow$DE~\cite{WMT}
for the NMT (Transformer) task.
A batch size of $128$ was used for CNNs, and $64$ was used for the rest of the
benchmarks.
We compare our results against data parallelism, expert-designed strategies,
and the strategies suggested by FlexFlow~\cite{flexflow}.


\newcommand\timecell[3]{\makebox[1.3cm][c]{#1}\makebox[1.3cm][c]{#2}\makebox[1.3cm][c]{#3}}
\begin{table*}[t!]
  \centering
	\caption{\label{tbl:running-times}Time taken by different algorithms to find
	efficient parallelization strategies. (Unit: \textit{mins:secs.msecs})}
  \reducespaceabovetbl
	\begin{tabular}{|r|p{3.9cm}|p{3.9cm}|p{3.9cm}|p{3.9cm}|}
		\hline 
		{\boldmath$p$} &
		\makebox[3.9cm][c]{\textbf{AlexNet}}\newline \timecell{BF}{FlexFlow}{Ours} &
		\makebox[3.9cm][c]{\textbf{InceptionV3}}\newline \timecell{BF}{FlexFlow}{Ours} &
		\makebox[3.9cm][c]{\textbf{RNNLM}}\newline \timecell{BF}{FlexFlow}{Ours} &
		\makebox[3.9cm][c]{\textbf{Transformer}}\newline \timecell{BF}{FlexFlow}{Ours}\\
		\hline
		4 & \timecell{0:00.234}{0:02.54}{0:00.226} & \timecell{OOM}{1:09.21}{0:14.398} &
		\timecell{0:00.070}{1:47.07}{0:00.057} & \timecell{OOM}{NA}{0:09.752} \\
		8 & \timecell{0:00.260}{0:02.77}{0:00.253} & \timecell{OOM}{2:26.51}{0:20.018} &
		\timecell{0:00.084}{2:44.29}{0:00.086} & \timecell{OOM}{NA}{0:28.798} \\
		16 & \timecell{0:00.303}{0:06.98}{0:00.295} & \timecell{OOM}{6:05.82}{0:39.791} &
		\timecell{0:00.109}{7:12.19}{0:00.069} & \timecell{OOM}{NA}{2:10.882} \\
		32 & \timecell{0:00.385}{0:07.92}{0:00.361} & \timecell{OOM}{15:37.96}{1:26.039} &
		\timecell{0:00.167}{11:08.22}{0:00.131} & \timecell{OOM}{NA}{9:13.022} \\
		64 & \timecell{0:00.485}{4:17.30}{0:00.475} & \timecell{OOM}{37:17.04}{3:16.253} &
		\timecell{0:00.271}{17:21.98}{0:00.215} & \timecell{OOM}{NA}{31:23.187} \\
		\hline
	\end{tabular}
	~\\
	$p$: Number of GPUs; BF: Breadth-First ordering; OOM: Out-Of-Memory; NA: Not
  Available.
  \reducespacebelowtbl
\end{table*}
\paragraph*{FlexFlow}
FlexFlow is a deep learning framework that automatically finds fast
parallelization strategies. It uses a general Markov Chain Monte Carlo
(MCMC) search algorithm to explore the search space.
When the search finishes, the framework returns the best strategy
it has discovered. As this approach is based on meta-heuristics, the framework
could get stuck in a local minima, returning a sub-optimal strategy.
An initial candidate from the search space needs to be provided to MCMC to begin
the search process, and the efficiency of the strategy found by FlexFlow might
also vary depending on the initial candidate. As suggested in~\cite[Section
6.2]{flexflow}, we use expert-designed strategies as the initial candidates in
our evaluation, so that FlexFlow can improve upon them.

\paragraph*{Expert strategies}
Expert-designed parallelization strategies were developed by domain experts
on a case-by-case basis. Since not
all DNNs have well-defined expert-designed strategies proposed, we chose the
ones that were the most relevant, as also used by the previous
works~\cite{opt-cnn,flexflow}.
For CNNs, \cite{owt} proposes using data parallelism for the
convolution layers, and switching to parameter parallelism for the fully-connected
layers.  This technique is referred to as \emph{``one weird trick'' (OWT)}.
We use this technique to
evaluate both AlexNet and InceptionV3.
For RNNs, a data+pipeline parallelism strategy was proposed in~\cite{gnmt},
where different layers of RNN are placed on different devices to achieve
pipeline parallelism, and each layer is replicated on the remaining devices for
data parallelism. We compare against this strategy for RNNLM.
For the Transformer model, we compare against the hybrid parallelism strategy
suggested in \cite{mesh-tf}, which primarily focuses on training large
Transformer models within the memory constraints, while also achieving good
parallel execution efficiency.
Since neither our technique, nor FlexFlow or various expert-designed strategies
used in our experiments perform any semantic changes to the DNN,
the final trained accuracy of all the strategies match the accuracy of the
original model.

\subsection{\label{subsec:expt-overhead}Runtime overhead}
In this subsection, we measure the time taken by different approaches to find
the best strategies for the four benchmarks. We compare the running
time of our approach that uses \sortnodes to order the vertices,
against breadth-first (BF) ordering (Subsection~\ref{subsec:bfs}), and FlexFlow
that uses meta-heuristics to find efficient strategies. 
We implemented our approach in a prototype tool written in Python, available at
https://github.com/baidu-research/PaSE.
The measurements were performed on a machine with Intel Xeon E5 (SandyBridge)
processor and 1080Ti GPUs.
Unlike our approach that uses analytical costs, FlexFlow microbenchmarks the
operators on GPUs and uses the execution results to find best strategies.
Table~\ref{tbl:running-times} shows the time taken by different
approaches to find the best strategies. For measuring the running times, as
suggested in~\cite[Section 6.2]{flexflow}, we stop FlexFlow's search algorithm
either when it is unable to improve the best discovered strategy for half the
search time, or when it has reached $250{,}000$ iterations.

As the computation graph of AlexNet is a simple path graph, sizes of both
$\bfsdependent{i}$ and $\dependent{i}$ are just one for different
vertices. Hence, both BF and \sortnodes ordering are able to efficiently compute
the best strategy in similar time.
However, for InceptionV3, BF ordering runs out of memory due to high node
degree of a few vertices as detailed in Subsection~\ref{subsec:eg-inception}.

For RNNLM, since an RNN operator (with LSTM cells) can be efficiently
represented in a single iteration space, we represent the complete RNN operator
(including the recurrent steps) as a single vertex in the computation graph. The
iteration space of an RNN operator is a five-dimensional space consisting of
layer, batch, sentence sequence (recurrent steps are captured by
this dimension), output, and hidden dimensions.
This is different from the way RNN is modelled in
FlexFlow. In FlexFlow, the recurrent dimension is unrolled (we use a unroll
factor of $40$, same as~\cite{flexflow}), and each iteration is represented as a
vertex in the graph.
By representing the whole RNN operator as a single vertex in our approach, in
addition to tremendously reducing the graph size, it also allows our approach to analyze
configurations that take advantage of inherent pipeline parallelism present
within an RNN operator. Configurations that split the `layer' and `sentence
sequence' dimensions capture intra-layer pipeline parallelism in a RNN
layer. 
With this representation, the computation graph of RNNLM reduces to a simple
path graph. Hence, both BF and \sortnodes orderings efficiently find
the best strategies within a second. 

Similar to InceptionV3, a Transformer model has a large number of sparse
vertices with a very few dense vertices. However, unlike InceptionV3, these
high degree vertices (such as the final output of encoder) have long live
ranges, that eliminate possible orderings that can reduce the dependent
sets as effectively as in InceptionV3. This causes \dpalg to take longer to
find the best strategy for Transformer.
As with InceptionV3, BF ordering fails to find the best strategy for
Transformer. We were unable to successfully implement
and analyze the Transformer model with FlexFlow for comparison.

\subsection{\label{subsec:expt-speedup}Comparison of performances of different
strategies}
\begin{figure*}[!th]
	\centering
  \subfloat[\label{fig:speedup-1080ti}Speedup on 1080Ti GPUs.]
  {\includegraphics[width=.95\linewidth,trim={0 0 0 .5cm},clip]{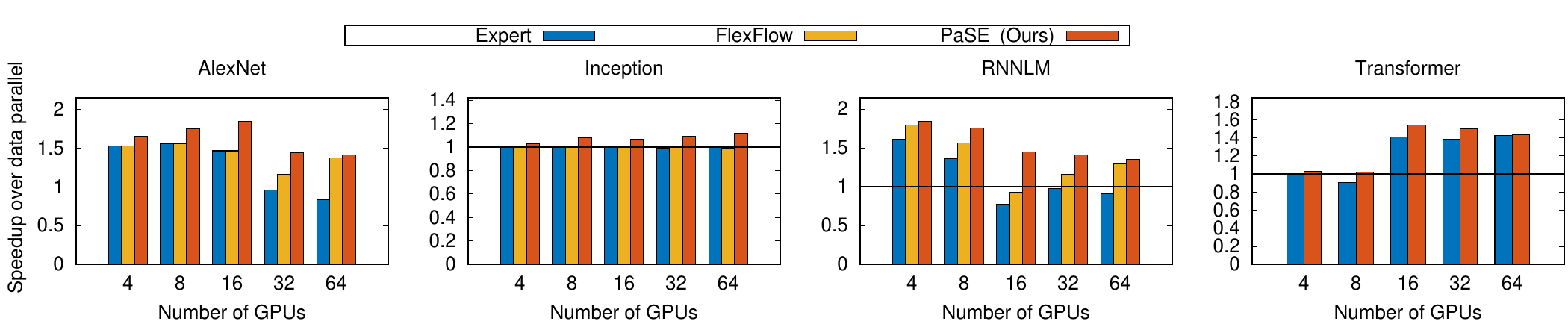}}
  \\
  \subfloat[\label{fig:speedup-2080ti}Speedup on 2080Ti GPUs.]
  {\includegraphics[width=.95\linewidth,trim={0 0 0 .5cm},clip]{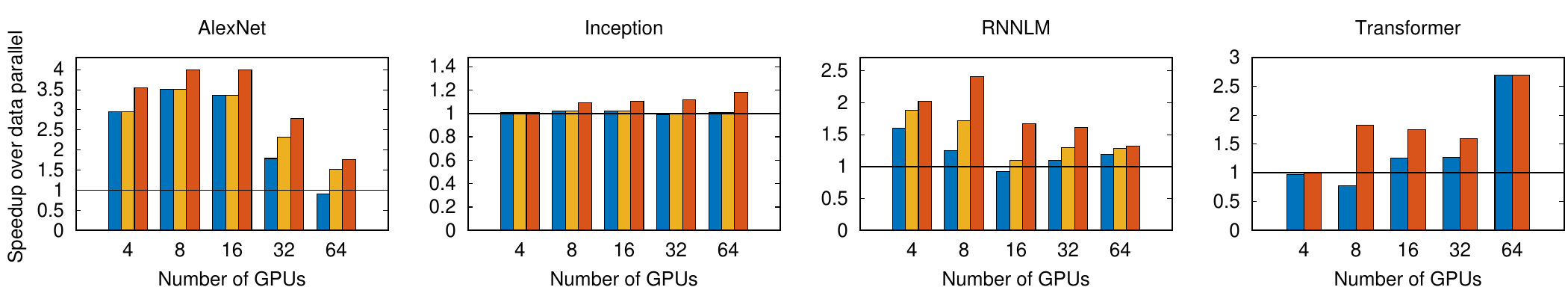}}
  \caption{\label{fig:speedup}Speedup achieved by various parallelism
  strategies over data parallelism.}
\end{figure*}
We compare the actual parallel training throughputs of the best
strategies proposed by our approach against data parallel, FlexFlow and expert
designed strategies.
The experiments were performed on varying number of GPUs ranging from $4$ (on
a single node) to $64$ (spread across $8$ nodes), incremented in powers of $2$.
The nodes are connected to each other using InfiniBand interconnection network.
We evaluated our results on the following two processing environments:
a)~a multi-node/multi-gpu system where each node contains 8 GeForce GTX 1080 Ti
GPUs (with $sm\_61$ compute capability) fully-connected using PCIe links;
b)~a multi-node/multi-gpu system where each node contains 8 GeForce RTX 2080 Ti
GPUs (with $sm\_75$ compute capability) fully-connected using PCIe links.
We implemented all the benchmarks in
Mesh-TensorFlow~\cite{mesh-tf} framework for evaluation.
Although our cost function (in Equation~\ref{eq:costfn}) ignores
accounting for certain optimizations such as 
inter-layer communication and computation overlap when computing costs for the
best strategies, all such feasible optimizations were allowed to be performed
by Mesh-TensorFlow in our experiments.
\fig~\ref{fig:speedup} shows the speedups achieved by various strategies 
over data parallelism on 1080Ti and 2080Ti systems.
On 1080Ti machines, the strategies proposed by our approach
achieve a speedup of up to $1.85\times$ over data parallelism. As shown
in \fig~\ref{fig:speedup-1080ti}, our strategies consistently perform
better than expert-designed strategies, and the strategies proposed by
FlexFlow. On 2080Ti machines, our strategies achieve up to $4\times$
speedup over data parallelism, and outperform both expert-designed strategies
and the strategies from FlexFlow.
2080Ti GPUs do not support peer-to-peer data access over PCIe
links, leading to poor hardware communication
efficiency, while having a higher computational peak than 1080Ti GPUs.
This leads to a very low machine balance (ratio between peak communication
bandwidth and peak GFLOPS). 
Thus, inefficiencies in parallelization strategies
are much more pronounced on 2080Ti nodes, allowing us
to achieve up to $4\times$ performance improvement over data parallelism.

\subsection{Analysis of computed strategies}
Table~\ref{tbl:strategies} shows the best strategies found by \dpalg for
training various DNNs on $p=32$ 1080Ti GPUs (spread across 4 nodes).

\begin{table}
  {\centering
  \caption{\label{tbl:strategies}Best strategies found by \dpalg for a system
  of 4 nodes, each consisting of 8 1080Ti GPUs ($p=32$).}
  \reducespaceabovetbl
  \setlength\tabcolsep{3.5pt}
  \begin{tabular}{l|lcc|l}
		\hline
    & \textbf{Layers} & \textbf{Dimensions} & \textbf{Configuration}
    & \textbf{Legend} \\
		\hline
    \rtxt{5}{\textbf{Alexnet}} &
    Conv 1--4 & $bchwnrs$ & $(32, 1, 1, 1, 1, 1, 1)$ &
    \mrow{9}{
      $b$: batch \\
      $h$: height\\
      $w$: width\\
      $c$: in-channel\\
      $n$: out-channel \\
      $r$: filter height \\
      $s$: filter width
    }\\
    & Conv 5 & $bchwnrs$ & $(16, 2, 1, 1, 1, 1, 1)$\\
    & FC 1, FC 3 & $bnc$ & $(1,4,8)$\\
    & FC 2 & $bnc$ & $(1,8,4)$\\
    & Softmax & $bn$ & $(1,4)$\\
    \cline{1-4}
    \rtxt{4}{\textbf{Inception}} &
    Modules A--D$^\dagger$ & $bchwnrs$ & $(32,1,1,1,1,1,1)$\\
    & Module E$^\dagger$ & $bchwnrs$ & $(16,1,1,1,2,1,1)$ \\
    & FC & $bnc$ & $(1,2,16)$\\
    & Softmax & $bn$ & $(1,2)$ \\
    \hline
    \rtxt{4}{\textbf{RNNLM}} &
    Embedding & $bsdv$ & $(1,1,1,32)$ &
    \mrow{9}{
      $b$: batch \\
      $s$: sequence len \\
      $d$: embed dim \\
      $e$: hidden dim \\
      $v$: vocab size \\
      $l$: RNN layers \\
      $h$: heads \\
      $c$: query channels \\
      $k$: kv channels
    }\\
    & LSTM & $lbsde$ & $(2,4,1,2,2)$ \\
    & FC & $bsvd$ & $(1,1,32,1)$ \\
    & Softmax & $bsv$ & $(1,1,32)$ \\
    \cline{1-4}
    \rtxt{5}{\textbf{Transformer}} &
    Embedding & $bsdv$ & $(1,1,1,16)$ \\
    & Multihead attn$^\dagger$ & $bshck$ & $(16,1,2,1,1)$ \\
    & Feed forward$^\dagger$ & $bsde$ & $(16, 1, 1, 2)$ \\
    & FC & $bsvd$ & $(1, 1, 16, 1)$ \\
    & Softmax & $bsv$ & $(1, 1, 16)$ \\
    \hline
  \end{tabular}}
  {\footnotesize $^\dagger$For simplicity, we report
    configurations at module level. In practice, they are further
    broken down into their constituent layers such as conv, FC, \etc}
    \reducespacebelowtbl
\end{table}

\emph{AlexNet} has five convolution layers, followed by three fully-connected layers.
On $32$ GPUs, our technique suggests to use data parallelism
for the first four convolution layer, and to split the in-channel
dimension of the last convolution layer into two.
For the first and third fully-connected layers,
the algorithm suggests to split the out-channel and in-channel dimensions by
4 and 8, respectively, while for the second fully-connected layer, it
suggests to split them by 8 and 4, respectively.
This alternating pattern effectively eliminates any
inter-layer communication among the fully-connected layers.
This differs from OWT~\cite{owt}, where only the out-channel dimension is
parallelized for fully-connected layers, leading to high volume of
all-gather communication between fully-connected layers.

\emph{Inception} network has a sequence of inception modules (A -- E) composed
of convolution layers, followed by a single fully-connected layer.
Our approach suggests to use
data parallelism for modules A--D, but for the module E,
the algorithm suggests a hybrid of data+parameter parallelism.
This is because as the modules get deeper, their output channels get larger,
and our algorithm finds pure data parallelism to be less
effective here.

\emph{RNNLM} is composed of an embedding layer, two layers of
LSTM cells, and a final projection layer, whose computations are dominated by
GEMM.
The embedding layer has a huge vocabulary dimension $v$, and a much smaller embedding
dimension $d$. \dpalg prefers fully splitting the vocabulary dimension
for the embedding and projection layers. For the LSTM cells, the algorithm suggests 
to fully split the LSTM layer dimension $l$ (thus utilizing intra-layer
pipeline parallelism), and partially split
the other three dimensions -- batch, hidden, and
output dimensions --  to varying degrees.

\emph{Transformer} is a non-recurrent self attention based NMT model.
A hybrid parallelism strategy was suggested in~\cite{mesh-tf}, where the batch
dimension of all the layers are split $m$-way, and model dimensions of
different layers -- vocabulary dimension, feed-forward hidden layer dimension, and
attention heads -- are split $n$-way.
Our approach suggests to use parameter parallelism for embedding and
softmax layers, and to use a hybrid data+parameter
parallelism for the remaining layers.

\section{\label{sec:limitations}Limitations and future work}
We discuss some of the limitations of our approach, and possible directions for
future work.
  
  \emph{Computational complexity} of our algorithm \dpalg is
    $O({|V|}^2K^{M+1})$, where $K=\max_{v\in V} |\allconfigs(v)|$, and
    $M=\max_{\vseqn{i}\in V} |\dependent{i}|$.
    DNN graphs are typically sparse allowing us to
    carefully order the vertices and efficiently compute
    parallelization strategies in practice. However, there do exist a few
    DNNs (such as DenseNet~\cite{densenet}) whose graphs are uniformly dense.
    No possible arrangement of vertices can effectively reduce the size $M$ for
    such graphs, leading to high runtime overhead for our algorithm.
  
  \emph{Inter-layer pipeline parallelism} is ignored in our approach. 
    Since DNNs typically do not have sufficient pipeline parallelism potential
    (without semantic changes), this does not severely affect our solutions.
    However, this prevents us from capturing the
    effects of overlapping computation and communication between different
    layers. (Computation / communication overlap within layers are
    accounted for in layer costs $t_l$.)
    Thus, it would be beneficial to incorporate pipeline parallelism into our
    formulation to improve its accuracy further.
  
  
  Although our method is applicable to \emph{heterogeneous architectures}, it
    does not explicitly include heterogeneity into the cost model. In case of
    heterogeneous systems, the peak FLOP and bandwidth, of the weakest
    computation node and communication link, respectively, are
    used to compute $t_l$ and $t_x$, as they form the primary bottlenecks.
    In future work, we plan to extend the model to include heterogeneity.
 
 For simplicity, we ignored several \emph{low level details} such as cache
 effects in our cost model.  This is not an inherent limitation of our approach
 as such, and in future work, we plan to fine-tune the cost model further by
 including these low level details to improve its accuracy.

\section{\label{sec:related}Related Work}
Data parallelism has been widely used as the standard technique to parallelize
DNNs. Data parallelism requires model parameters to be fully replicated on all
devices.
This typically leads to poor performance and
scalability for large models. Some previous works
\cite{zero,automatic_weight_update} have tried to address this by
storing the parameters sharded across different devices, while
still using data parallelism for computation. This leads to additional
communication.
Although these techniques propose methods to
efficiently perform these communications, the minimum volume of data
that needs to be communicated remains the same,
leading to
a fundamental bottleneck.  In contrast, our method splits computation of each
layer along different dimensions, minimizing the actual communication volume.
Memory and communication optimizations proposed by
\cite{automatic_weight_update, zero} are thus orthogonal to ours, and can be
independently applied on top of ours to further improve the performance.

\emph{One weird trick} (OWT) was introduced in \cite{owt} to parallelize CNNs,
where data parallelism is used for convolutional layers, and parameter
parallelism is used for the rest. This trick, while
applicable only for CNNs, works reasonably well in practice. However, as
evident from \fig~\ref{fig:speedup}, even better performance can be achieved by
a more sophisticated hybrid parallelism.
A DP based approach to automatically find efficient strategies for
CNNs was presented in~\cite{opt-cnn}. The method exploits the fact that 
CNNs typically have nodes with single in-/out-edges, and
computes efficient strategies by reducing the graph through \emph{node and edge
eliminations}.
However, this technique fails on other tasks such as LM and NMT whose graphs do
not have this special property. 
In contrast, our method is not limited to CNNs, and can find efficient
strategies for various types of networks like RNNs and Transformers within
a few minutes.
Additionally, we define our parallelization configuration to split
any dimension in the iteration space, while in \cite{opt-cnn}, only the output
tensor dimensions are split. This heavily restricts the search space,
since some of the dimensions are not considered as possible choices for
parallelization.
\emph{Tofu}~\cite{tofu} also uses the same DP formulation proposed in
\cite{opt-cnn} to find the best strategies to parallelize fine-grained dataflow
graphs, and
thus suffers from the same limitation as \cite{opt-cnn},
preventing them from being able to handle models such as Transformer, whose
graphs do not have a linear structure.

\emph{FlexFlow}~\cite{flexflow} uses a general Markov Chain Monte Carlo (MCMC)
meta-heuristic search algorithm to explore the search space to discover the
best strategy. The strategy returned by their framework need not necessarily be
optimal. While their method takes the whole search space into consideration,
our method ignores inter-layer pipeline parallelism.  In return, our method is
able to find an efficient strategy for various DNNs much faster than FlexFlow,
and is not subject to the limitation of getting stuck at a local minima.
REINFORCE~\cite{reinforce} and \cite{hierarchical-device-placement} use
machine learning to find efficient device placement for various layers to
achieve efficient pipeline parallelism.  They ignore data
and parameter parallelism in their search process. Further, they
require multiple GPUs and take several hours to find an efficient strategy.
%
\cite{diesel} and \cite{tc} use polyhedral compilation techniques to optimize
execution of individual DNN operators on a single GPU.
		These techniques can be orthogonally used with ours to further
		improve the performance within each GPU.

\emph{PipeDream} \cite{pipedream} allows for semantic
changes to the model to improve parallel training times. 
In our approach, we do not consider semantic modifications.
While our technique heavily relies on parameter parallelism, this is completely
ignored in~\cite{pipedream}. 
Thus, we find their approaches to be complementary to ours: the computation
graph can be first split into multiple stages using the formulation proposed in
\cite{pipedream} to achieve inter-batch pipeline parallelism, and the subgraphs
from each stage can be further parallelized with data+parameter parallelism
using our approach.

Several expert-designed strategies \cite{owt, gnmt} have been proposed for
different networks based on domain specific knowledge. Each network
has to be individually analyzed manually to come up with an efficient strategy.
Further, these strategies need not be necessarily optimal. The method proposed
in this paper automates this process, and can point the expert towards the
right direction for parallelization.
While the focus of this paper is to find the best parallelization strategies
for DNNs, frameworks such as Mesh-TensorFlow~\cite{mesh-tf} and
GShard~\cite{gshard} enable automatically converting these user-specified
strategies into efficient parallel programs. 

\section{\label{sec:conclusion}Conclusion}
In this paper, we presented a method to automatically find efficient
parallelization
strategies for DNNs. We proposed a recurrence formulation to compute
the minimum cost of a computation graph, and presented a technique to
efficiently compute the best parallelization strategies within a few minutes.
We evaluated our results against data parallelism, expert
designed strategies, and the strategies proposed by a deep learning framework,
FlexFlow. Results show that the strategies proposed by our method
outperform the standard data parallelism by a factor of up to $1.85\times$ and
$4\times$ on multi-node systems with 1080Ti and 2080Ti GPUs, respectively. In
addition, strategies from our method perform better than expert-designed
strategies, and the ones proposed by FlexFlow.

\section*{Acknowledgment}
We thank Kenneth Ward Church, Prashant Singh Rawat, and the anonymous reviewers
for their valuable feedback and suggestions that helped improve the paper.

\bibliographystyle{IEEEtran}
\bibliography{PaSE_ipdps2021}

{\footnotesize
\appendix
\subsection{\label{subsec:optimality}Optimality of \dpalg}
\optimality*
\begin{IEEEproof}
  From Equation~(\ref{eq:recurrence}), we have,
	\begin{align}
    &\reccostfn(|V|, \emptyset) \nonumber\\
      ={}& \min_{C\in \allconfigs(\vseqn{|V|})}~
      \nodefn(|V|, \setn{(\vseqn{|V|},C)}) + 
      \sum_{\substack{\connected{j}\in \\ \sg{|V|}}}
    \reccostfn(j, \setn{(\vseqn{|V|},C)}) \nonumber\\
      ={}& \min_{\phi\in \Phi} \sum_{\vseqn{i}\in V}
    \nodefn(i, \phi) \label{ln:hset}\\
        ={}& \min_{\phi\in \Phi} \sum_{\vseqn{i}\in V} \bigg[ t_l(\vseqn{i},
      \phi, r)+
      \sum_{\vseqn{j}\in N(\vseqn{i})\cap \vseq_{>i}} r\times t_x(\vseqn{i}, \vseqn{j},
			\phi)\bigg]\nonumber\\
        ={}& \min_{\phi\in \Phi} \sum_{\vseqn{i}\in V} \bigg[t_l(\vseqn{i},
      \phi, r)\bigg] + \sum_{(\vseqn{i}, \vseqn{j})\in E} \bigg[r\times
      t_x(\vseqn{i}, \vseqn{j}, \phi)\bigg]\label{ln:nset}\\
         ={}& \min_{\phi\in \Phi} \costfn(G, \phi)\nonumber
	\end{align}

  Equality~(\ref{ln:hset}) is due to the fact that for any vertex $\vseqn{i}\in
  V$, $\connected{i} = (\bigcup_{U\in \sg{i}} U)\cup \setn{\vseqn{i}}$, and the
  connected \mbox{sub-components} in $\sg{i}$ are pairwise disjoint, \ie for $U_1,
  U_2\in \sg{i}$, $U_1\not=U_2 \implies U_1\cap U_2 = \emptyset$. Further,
  a computation graph $G$ is weakly connected. Thus, $\connected{|V|} = V$.

  Equality~(\ref{ln:nset}) is due to the fact that the union of the pairwise
  disjoint sets $\set{\{\vseqn{i}, \vseqn{j}\}} {\vseqn{j}\in N(\vseqn{i})\cap
  \vseq_{>i}}$ is the set of edges of (undirected) $G=(V,E')$, where $E'=\set{\setn{u,
  v}}{(u,v)\in E}$.
\end{IEEEproof}

\subsection{\label{subsec:correctness}Correctness of \sortnodes}
\correctness*
\begin{IEEEproof}
	We will show this by induction.
  Let $\vseq$ be the sequence generated by \sortnodes.
  We define the dependent set of a vertex $\vseqn{j}$ restricted to the first
  $k$ vertices in the sequence,
  $\dependent{j}_{|k} = N(\connected{j} \cap \vseq_{\le k} \cup
  \setn{\vseqn{j}}) \cap \vseq_{>\min(j, k)}$.
  Clearly, for any $j\le k$, $\dependent{j}_{|k} = \dependent{j}$.
  We will show that at the end of any iteration
  $i$ in \sortnodes, the invariant $\vseqn{j}.d = \dependent{j}_{|i}$ holds.
  This will prove that at the end of the algorithm, for any $j$, $\vseqn{j}.d
  = \dependent{j}_{||V|} = \dependent{j}$.
  \paragraph*{Induction base}
  The invariant trivially holds just before the first iteration (where
  $i=0$) due to the initialization in Line~\ref{ln:sort-init-d}.

  \paragraph*{Induction step}
  As a hypothesis, consider that the invariant is true at the end of an
  iteration $i-1$.
  Let $\vseqn{i}$ be the vertex chosen at iteration $i$ in
  Line~\ref{ln:sort-set-v}.
  For any $j$ \st $\vseqn{i}\not\in \connected{j}$, 
  $\dependent{j}_{|i} = N(\connected{j} \cap \vseq_{\le i} \cup
  \setn{\vseqn{j}}) \cap \vseq_{>i} = N(\connected{j} \cap \vseq_{< i} \cup
  \setn{\vseqn{j}}) \cap \vseq_{\ge i} = \dependent{j}_{|i-1}$.
  For any $j$ \st $\vseqn{i}\in \connected{j}$ (\ie if $\vseqn{j}\in
  \vseqn{i}.d$), 
  \begin{align}
    \dependent{j}_{|i} ={}& N(\connected{j} \cap \vseq_{\le i} \cup
    \setn{\vseqn{j}}) \cap \vseq_{>i} \nonumber\\
    ={}& N((\connected{j}\cup \connected{i}) \cap\vseq_{\le i} \cup
    \setn{\vseqn{j}}) \cap \vseq_{>i} \label{ln:expand}\\
    ={}& N((\connected{j}\cup \connected{i}) \cap\vseq_{<i} \cup
    \setn{\vseqn{i},\vseqn{j}}) \cap \vseq_{>i} \nonumber\\
    ={}& N((\connected{j}\cap \vseq_{<i} \cup \setn{\vseqn{j}}) \nonumber\\
    &\cup (\connected{i}\cap \vseq_{<i} \cup \setn{\vseqn{i}})) \cap
    \vseq_{>i-1} - \setn{\vseqn{i}}
    \nonumber\\
    ={}& \dependent{j}_{|i-1} \cup \dependent{i}_{|i-1} - \setn{\vseqn{i}}\nonumber
  \end{align}
  Equality~(\ref{ln:expand}) is due to the fact that $\connected{i}\subseteq
  \connected{j}$.
  The update in Line~\ref{ln:sort-vd} in Fig.~\ref{alg:sortnodes} indeed
  performs this exact operation, making sure that the invariant is correctly
  maintained at the end of iteration $i$. Thus, at the end of $|V|$ iterations,
  $\vseqn{j}.d = \dependent{j}$ for any $\vseqn{j}\in V$.
\end{IEEEproof}

}

\end{document}